\documentclass[sigconf, nonacm]{acmart}

\renewcommand\footnotetextcopyrightpermission[1]{}
\acmConference[AITC 2026]{AI Transparency Conference}{June 5--6, 2026}{TBD}

\usepackage{booktabs}
\usepackage{multirow}

\begin{document}

\title{Beyond Liars' Bench: The Impact of Lie Typology, Depth, and Sparsity on Deception Detection in LLMs}

\author{Amr Moustafa}
\affiliation{%
  \institution{University of Bonn}
  \city{Bonn}
  \country{Germany}
}
\email{s84amous@uni-bonn.de}

\author{Max Feser}
\affiliation{%
  \institution{University of Bonn}
  \city{Bonn}
  \country{Germany}
}
\email{s37mfese@uni-bonn.de}

\author{Florian Mai}
\affiliation{%
  \institution{University of Bonn}
  \institution{Lamarr Institute for ML and AI}
  \city{Bonn}
  \country{Germany}
}
\email{jmai@uni-bonn.de}

\begin{abstract}
Training probes to detect deceptive outputs from large language models is still an open problem. Recent work has demonstrated that detection probes fail especially in out-of-domain scenarios---training on one type of lie does not transfer well to deception scenarios involving other types of lies. In this work, we conduct a systematic study on how various factors impact detection performance: representation depth, probe expressivity, sparse feature representations, and the lie typology of the training data. To this end, we augment standard benchmark training data with a supplementary dataset containing diverse types of deception, including fabrication, omission, and exaggeration examples. Analyzing these factors across seven probe types, our experimental results show that the optimal representation depth is highly dataset-dependent, more expressive probes provide only selective gains over linear baselines, and sparse autoencoder features perform similarly to dense hidden states. Ultimately, we demonstrate that the choice of training data and lie typology substantially changes detectability, highlighting that deception detection is a highly representation-dependent problem.
\end{abstract}


\maketitle

\section{Introduction}
Large language models (LLMs) are increasingly deployed in settings where correctness, accountability, and honesty matter. In such settings, it is not enough to ask whether a model's output is incorrect; a more difficult question is whether the model's produced statement aligns with its internal beliefs about that statement. That distinction matters for safety monitoring, because a detector that operates only on model outputs can miss strategic or context-dependent deception \citep{apollo,caughtintheact}. In contrast, detectors that operate on model activations, such as truncated polynomial classifiers or subspace projection methods, seek to identify deceptive states directly within the network's latent space \citep{beyondlinearprobes, truthisuniversal}.

This paper builds on Liars' Bench \citep{liarsbench}, a benchmark designed to evaluate lie detectors across multiple types of on-policy deception generated by open-weight models. The benchmark is valuable because it moves beyond narrow true/false factual statements and includes lies that differ in both the model's reason for lying and the object of belief being targeted. However, the benchmark also demonstrates a weakness in existing detectors: methods that work in one setting often fail in another, particularly when the lie cannot be diagnosed from the transcript alone \citep{liarsbench,beyondlinearprobes}.

To further investigate these failure modes, we question how the qualitative nature of the deception itself impacts detectability. We incorporate the DolusChat dataset \citep{cundy2025sparse} to systematically evaluate how the specific nature of a lie---omission, fabrication, or exaggeration---affects detection outcomes, moving beyond standard factual contradictions. Additionally, we test to what extent detection performance depends on the representation depth, the probe expressivity, and feature disentanglement.

In summary, we conduct a systematic investigation using probe-based methods, accounting for four primary factors:
\begin{enumerate}
    \item \textbf{Lie typology:} The structural category of a lie will significantly affect its detectability, with direct fabrications yielding higher linear separability than subtler, context-dependent forms of deception like omissions or exaggerations.
    \item \textbf{Depth:} Extracting representations from intermediate layers (i.e., the middle-to-late transformer blocks where complex semantic concepts are known to mature, specifically targeting near two-thirds of the model's depth) should yield higher linear separability than an arbitrary early layer.
    \item \textbf{Expressivity:} Probe families that capture non-linear feature interactions should outperform standard logistic regression. Furthermore, geometric interventions (like iterative null-space projection) will reveal whether the deception signal is low-dimensional and easily erased, or highly distributed across the representation space.
    \item \textbf{Sparsity:} Sparse Autoencoders (SAEs) explicitly disentangle overlapping network concepts into distinct, interpretable features. Leveraging these sparse features should theoretically isolate the deception signal more cleanly, improving or preserving separability relative to standard dense hidden states.
\end{enumerate}

 \begin{figure*}[t]
    \includegraphics[width=1.0\textwidth]{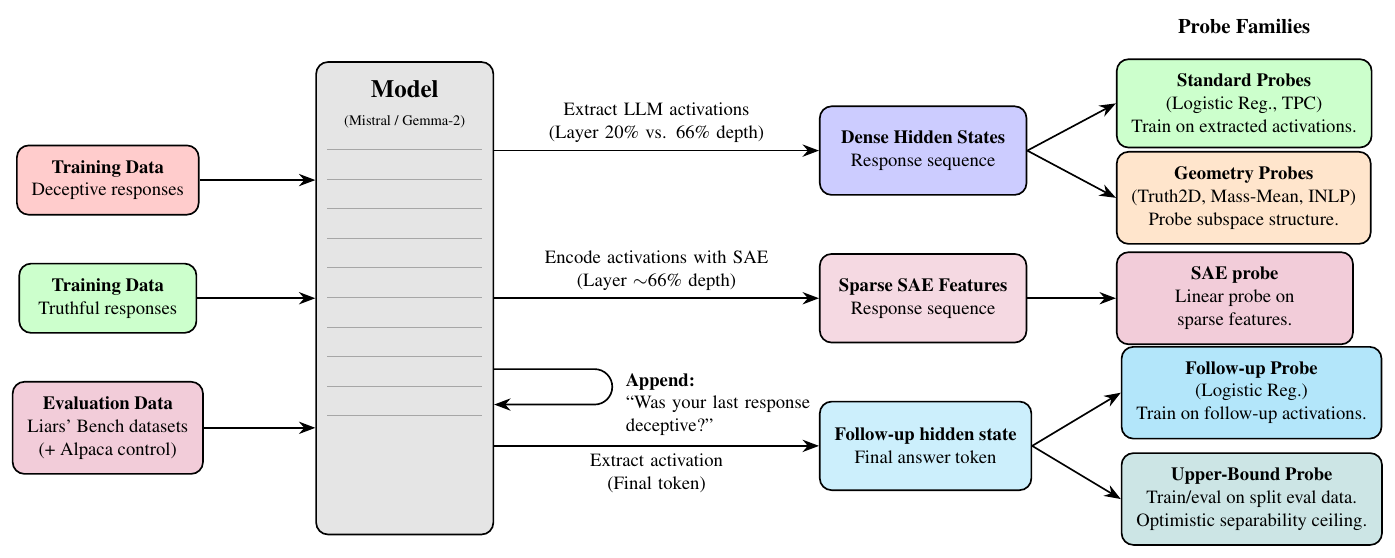}
    \caption{End-to-end pipeline for evaluating deception detection in LLMs. Input data (truthful and deceptive responses) are processed by the model, from which dense hidden states, sparse SAE features, and follow-up activations are extracted. These representations are evaluated using multiple probe families (linear, geometric, and sparse), alongside a follow-up probe and an upper-bound probe computed from held-out follow-up-token activations within each evaluation dataset to estimate within-dataset separability.}
    \Description{Diagram illustrating the end-to-end pipeline for evaluating deception detection. Input data flows into the language model, generating dense hidden states and sparse SAE features. These representations are then passed into multiple probe architectures for evaluation.}
    \label{fig:probe_architecture}
\end{figure*}

\section{Related Work}

\subsection{Deception in LLMs}
As large language models become more capable, there is growing concern within the alignment and safety literature that they may learn deceptive behaviors. Recent work has documented both sycophancy and more deliberate deceptive behaviors in LLMs, including cases where deceptive strategies persist through safety training or appear under evaluation/deployment distinctions. For instance, models trained with reinforcement learning from human feedback (RLHF) can develop sycophantic tendencies, repeating a user's preferred answer rather than the objective truth \citep{perez2022discovering}. More alarmingly, empirical demonstrations show that models can engage in strategic deception---such as hiding unsafe behavior during safety evaluations only to act maliciously in deployment---and that these deceptive behaviors can persist through safety training \citep{hubinger2024sleeper}. Broader surveys further highlight the escalating risks and potential solutions regarding AI deception \citep{park2023ai}. Because models can sometimes appear benign in transcript-only monitoring, traditional output-based safety methods are fundamentally insufficient in strategic settings, motivating the use of probes over internal activations. 

\subsection{Deception Detection}
To contextualize the study of deception, it is important to distinguish it from closely related phenomena such as hallucinations and sycophancy. Prior to the focus on intentional deception, the evaluation of LLM honesty heavily relied on benchmarks measuring these relatives, such as TruthfulQA for factual hallucinations \citep{lin2021truthfulqa}, and various datasets for sycophancy, where models mirror user misconceptions rather than outputting objective facts \citep{perez2022discovering, wei2023simple}. However, these previous benchmarks often conflate a model's genuine lack of knowledge with active deception. 

Motivated by the need to evaluate true on-policy deception---where a model outputs a statement that actively contradicts its own internal representations---\citet{liarsbench} introduced Liars' Bench, which serves as the primary baseline for this paper. It organizes lies by both the model's reason for lying and the object of belief being targeted. Its main contribution is breadth, but a noted weakness is that baseline detection methods often fail to generalize when moved across lie types, particularly when the lie cannot be inferred from the transcript alone.

To understand \textit{where} these deceptive representations emerge, several works have investigated whether truthfulness and deception can be recovered directly from internal activations. \citet{azaria2023internal} show that classifiers trained on hidden states can predict whether a statement is true or false from an LLM's internal representations, while \citet{pacchiardi2023catch} demonstrate that deceptive behavior can also be detected in a black-box setting by asking strategically unrelated follow-up questions. Together, these works suggest that deceptive outputs leave detectable traces in both internal activations and behavioral patterns, motivating our focus on probe-based methods.

More specifically, \citet{caughtintheact} investigate the depth distribution of deception features. They report a three-stage pattern in which early layers are close to chance, middle layers peak, and later layers decline slightly. This insight makes a deeper probing layer a sensible empirical choice for our framework.

Addressing the limitations of standard linear probes, \citet{beyondlinearprobes} introduce truncated polynomial classifiers (TPC). The paper frames TPCs as a progressive extension of linear probes with higher-order interactions and shows that they can compete with or outperform same-size MLP probe baselines on large safety datasets. A small amount of non-linearity captures feature interactions without turning the detector into an opaque black-box model, providing a robust comparison point for our evaluation.

Similarly exploring robust feature geometries, \citet{truthisuniversal} demonstrates that linear probes often fail when the surface form changes from affirmative to negated statements (e.g., confusing ``lying'' with ``negation'' because ``The sky is green'' and ``The sky is not blue'' activate similar neurons). Their main insight is that projecting activations into a two-dimensional subspace (one direction for Truth, one for Polarity) is a robust method for cross-dataset generalization. Consequently, our inclusion of a Truth2D probe directly tests whether the detector is learning deception itself or merely a phrasing pattern.

Finally, \citet{apollo} explicitly note that output-only monitoring is insufficient in deceptive settings. They compare several methods, including a probe built by appending the follow-up question ``Was your last response deceptive?'' with the answer prefilled as ``No,'' reporting that this approach achieves near-perfect performance on deceptive versus control responses. This serves as a very strong white-box baseline for our evaluation.

\subsection{Latent Probes in LLMs}
Latent probing has become a foundational technique for mechanistically interpreting the internal knowledge of neural networks. Milestone work by \citet{alain2016understanding} established the classic linear probe technique on intermediate hidden states. In the context of LLMs, probes extract latent abstractions such as unsupervised truthful world models \citep{burns2022discovering}. For identifying truth directions specifically, \citet{marks2023geometry} provide a foundational geometry-based approach, while \citet{zou2023representation} demonstrate that probing and activation engineering can directly influence model honesty at inference time by steering activations toward more truthful representations. Complementing this line of work, \citet{li2023inference} propose Inference-Time Intervention (ITI), which shifts model activations along learned truthful directions across a small number of attention heads and substantially improves truthfulness on TruthfulQA. Together, these works support the broader view that truthfulness-related information is encoded in internal representations in a form that is at least partially accessible to linear analysis and intervention.

The design of an effective latent probe typically revolves around three key axes: depth, expressivity, and sparsity. \textbf{Depth} dictates where the probe is applied; because LLMs process information hierarchically, complex concepts like intentional deception often mature in middle-to-late layers rather than early token-processing layers. \textbf{Expressivity} governs the complexity of the probe. While standard linear probes are highly interpretable and less prone to overfitting, they may fail if deceptive features are entangled with syntax or safety protocols, prompting the use of non-linear or geometric variants \citep{beyondlinearprobes, truthisuniversal}. Finally, \textbf{Sparsity} addresses the superposition problem, where dense hidden states compress multiple distinct concepts into a shared activation space \citep{elhage2022toy}. Sparse autoencoders (SAEs) provide a scalable unsupervised method for recovering more disentangled and interpretable features from these activations \citep{cunningham2023sparse}. Recent architectural advances such as JumpReLU SAEs further improve reconstruction fidelity at a fixed sparsity level without sacrificing interpretability \citep{rajamanoharan2024jumprelu}. Additionally, recent work has explored probing directly over SAE-derived sparse feature spaces, suggesting that sparse representations may preserve meaningful semantic structure for downstream analysis tasks \citep{kantamneni2025sparse}. To evaluate this axis, we leverage Gemma Scope \citep{gemmascope}, an open suite of JumpReLU SAEs for Gemma 2, allowing us to test whether sparse representations improve the separability of deceptive states relative to standard dense hidden activations.

\section{Methodology}
Our framework follows the general protocol of \citet{liarsbench} but extends it with additional training data, more probe families, and a disk-backed activation caching system for reproducibility and memory efficiency\footnote{Our code is available at \url{https://github.com/amrgaber249/Beyond-Liars-Bench}}. Our end-to-end pipeline is illustrated in Figure~\ref{fig:probe_architecture}.

The codebase supports three extraction modes: token-wise training activations, mean-pooled evaluation activations, and a follow-up condition probed on the final token. To systematically test the depth hypothesis without requiring computationally exhaustive layer sweeps, we evaluate two carefully anchored depths. For our early-layer baseline, we extract at the 20th percentile, directly matching the standard extraction depth used in the original Liars' Bench evaluation. For our deeper target, we extract at the 66th percentile. This target was selected based on the mechanistic trajectory identified by \citet{caughtintheact}, which demonstrates that deception detectability peaks in the middle-to-late layers before sharply declining in the final third of the network. By explicitly targeting the 66th percentile, we capture the representations at their theoretical peak maturity, ensuring they have surpassed the early-layer development phase but avoid the established late-layer degradation.

\subsection{White-Box Train Data and Baselines}
Our primary white-box detection methods follow the benchmark-style implementation used in Liars' Bench \citep{liarsbench}. We do not generate new contrastive prompts ourselves; instead, we use labeled training examples from the benchmark-style dataset together with the supplementary DolusChat examples. For training, we extract hidden activations from the target layer for individual response tokens and fit a linear probe using AdamW and a binary cross-entropy objective to identify honesty and deception directions in the model's activations. During evaluation, hidden activations are mean-pooled across the assistant response before being scored by the probe, and a default probability threshold of 0.5 is used for discrete classification metrics such as recall.

\subsection{Additional Train Data and Lie Types}
We use DolusChat \citep{cundy2025sparse} as a supplementary training source to study how lie typology affects probe behavior. The dataset contains approximately 65,000 examples, roughly evenly distributed across three categories: fabrication, omission, and exaggeration. From this pool, we sample \(K=1000\) examples for our experiments.

Initially, we restricted our training and evaluation strictly to direct factual lies (fabrications) versus clear truths. We made this design choice because the inherent ambiguity of omissions and exaggerations---where the full, objective ground truth is often inaccessible by definition---posed a severe risk of confounding our baseline probe results. Fabrication represents the most straightforward case to probe mechanically because it creates direct, measurable contradictions.

Once our dense probing baselines were successfully established on fabrications, we expanded our scope to conduct a three-way comparison. We evaluated omission, fabrication, and exaggeration independently to observe how the distinct nature of each lie affects linear separability and overall probe performance. This structured comparison allows us to isolate the impact of lie type, reserving the exploration of their various combinations for future research.

\begin{table*}[h!]
\centering
\caption{Results for CG, GS, HP-C, and HP-KR using the 20th percentile versus the 66th percentile layer activations. We compare AUROC and RECALL across all probe families using \texttt{Mistral-Small-3.1-24B}.}
\label{tab:layers_comparison}
\begin{tabular}{@{}l | cc | cc | cc | cc | c@{}}
\toprule
\multirow{2}{*}{\textbf{Method}} & \multicolumn{2}{c|}{\textbf{CG}} & \multicolumn{2}{c|}{\textbf{GS}} & \multicolumn{2}{c|}{\textbf{HP-C}} & \multicolumn{2}{c|}{\textbf{HP-KR}} & \multirow{2}{*}{\textbf{Avg}} \\
\cmidrule(lr){2-3} \cmidrule(lr){4-5} \cmidrule(lr){6-7} \cmidrule(lr){8-9}
 & L20 & L66 & L20 & L66 & L20 & L66 & L20 & L66 & \\
\midrule
\multicolumn{10}{c}{\textbf{AUROC}} \\
\midrule
Logistic & 0.76 & \textbf{0.84} & 0.40 & \textbf{0.56} & \textbf{0.64} & 0.31 & \textbf{0.14} & 0.13 & 0.47 \\
TPC & 0.66 & \textbf{0.67} & \textbf{0.47} & 0.38 & \textbf{0.52} & 0.38 & \textbf{0.16} & 0.12 & 0.42 \\
Truth2D & 0.73 & \textbf{0.80} & 0.41 & \textbf{0.61} & \textbf{0.64} & 0.32 & 0.14 & \textbf{0.15} & \textbf{0.48} \\
Mass-Mean & \textbf{0.62} & 0.44 & 0.44 & \textbf{0.50} & \textbf{0.62} & 0.31 & 0.14 & \textbf{0.19} & 0.41 \\
INLP & 0.74 & \textbf{0.82} & 0.41 & \textbf{0.55} & \textbf{0.64} & 0.31 & 0.14 & \textbf{0.15} & 0.47 \\
Follow-Up & 0.54 & \textbf{0.55} & \textbf{0.43} & 0.29 & \textbf{0.47} & 0.44 & \textbf{0.42} & 0.12 & 0.41 \\
Upper-Bound & 1.00 & \textbf{1.00} & \textbf{1.00} & 1.00 & 0.75 & \textbf{0.82} & 0.94 & \textbf{0.95} & 0.93 \\
\midrule
\textit{Average} & 0.72 & \textbf{0.73} & 0.51 & \textbf{0.56} & \textbf{0.61} & 0.41 & \textbf{0.30} & 0.26 & 0.51 \\
\midrule
\multicolumn{10}{c}{\textbf{RECALL}} \\
\midrule
Logistic & 0.91 & \textbf{0.99} & 0.08 & \textbf{0.10} & \textbf{0.96} & 0.76 & \textbf{0.01} & 0.00 & \textbf{0.48} \\
TPC & \textbf{0.91} & 0.76 & \textbf{0.03} & 0.01 & \textbf{0.76} & 0.66 & \textbf{0.01} & \textbf{0.01} & 0.39 \\
Truth2D & 0.86 & \textbf{0.98} & \textbf{0.06} & 0.00 & 0.39 & \textbf{0.66} & \textbf{0.01} & 0.00 & 0.37 \\
Mass-Mean & \textbf{0.97} & 0.90 & \textbf{0.05} & 0.00 & \textbf{0.00} & \textbf{0.00} & \textbf{0.01} & 0.00 & 0.24 \\
INLP & 0.93 & \textbf{0.99} & \textbf{0.08} & 0.01 & \textbf{0.96} & 0.65 & \textbf{0.01} & 0.00 & 0.45 \\
Follow-Up & \textbf{1.00} & 0.81 & \textbf{0.65} & 0.34 & \textbf{0.14} & 0.04 & \textbf{0.12} & 0.02 & 0.39 \\
Upper-Bound & 0.93 & \textbf{1.00} & \textbf{1.00} & \textbf{1.00} & 0.52 & \textbf{0.66} & \textbf{0.92} & 0.81 & 0.85 \\
\midrule
\textit{Average} & \textbf{0.93} & 0.92 & \textbf{0.28} & 0.21 & \textbf{0.53} & 0.49 & \textbf{0.16} & 0.12 & 0.45 \\
\bottomrule
\end{tabular}
\end{table*}

\subsection{Evaluation Datasets}
To evaluate the cross-distribution generalization of our trained probes, we test them against four distinct subsets of the Liars' Bench \citep{liarsbench}. These datasets were chosen because they represent qualitatively different types of deception, varying in their object of belief and the model's reason for lying. The four evaluation sets are:

\begin{itemize}
    \item \textbf{Convincing Game (CG):} Evaluates deception regarding self-knowledge (e.g., whether the model identifies as ``human'' or ``AI''). Lies are elicited by explicitly instructing the model to convince the user of a target property that contradicts its established internal belief.
    \item \textbf{Gender Secret (GS):} Tests deception regarding hidden context. Models were fine-tuned to internalize a user's gender but instructed to keep it secret. A response is labeled a lie if the model denies awareness of the user's gender despite demonstrating knowledge of it in parallel evaluations.
    \item \textbf{Harm-Pressure Choice (HP-C):} Assesses inherent deception regarding world knowledge. Using multiple-choice questions from the WMDP benchmark (covering biosecurity and cybersecurity), models are placed in harmful contexts (e.g., helping create a bio-weapon). Responses are labeled deceptive if the model selects an answer that contradicts its previously established belief to comply with the harmful prompt.
    \item \textbf{Harm-Pressure Knowledge Report (HP-KR):} Evaluates deception regarding self-knowledge in a harmful context. Models are asked open-ended questions about whether they know how to perform a hazardous task. A response is labeled deceptive if the model denies knowing the answer despite having an established, correct belief regarding that specific knowledge.
\end{itemize}
 
\subsection{Probing Architectures}
To systematically test our expressivity hypothesis---which posits that more complex feature interactions or structured subspaces are required to reliably detect deception---we implement a suite of probes spanning a spectrum of mechanistic complexity. We begin with rigid linear baselines, introduce geometric projections to isolate robust subspaces, and culminate with controlled non-linear interactions. Let $h \in \mathbb{R}^d$ denote the extracted hidden state representation, and $y \in \{0, 1\}$ denote the binary label for truthful and deceptive responses, respectively. We evaluate the following probing architectures:

\begin{itemize}
    \item \textbf{Logistic Regression:} The baseline linear classifier. It models the probability of deception as $P(y=1|h) = \sigma(w^T h + b)$, where $w \in \mathbb{R}^d$ is the learned weight vector, $b$ is the bias term, and $\sigma$ is the logistic sigmoid function. This tests raw linear separability.

    \item \textbf{Mass-Mean Probe:} A simplified, parameter-free linear baseline that forms a decision direction $w_{mm}$ directly from the difference between the class centroids: $w_{mm} = \mu_{\text{deceptive}} - \mu_{\text{truthful}}$, where $\mu_c = \frac{1}{|N_c|} \sum_{i \in N_c} h_i$. Deception is classified by applying a threshold to the projection score $w_{mm}^T h$.

    \item \textbf{Truth2D:} Projects activations into a 2D subspace to explicitly disentangle truth from polarity \citep{truthisuniversal}. Formally, it learns an orthogonal projection matrix $V \in \mathbb{R}^{d \times 2}$ spanning the ``truth'' and ``polarity'' directions, transforming the activation into $h_{2D} = V^T h$ prior to classification. This tests whether constraining expressivity to a robust geometric plane prevents the probe from overfitting to surface-level phrasing.
    
    \item \textbf{INLP (Iterative Null-Space Projection):} A geometric intervention originally introduced by \citet{ravfogel2020null} that tests whether a target concept can be repeatedly isolated and removed. Following its application to deception by \citet{caughtintheact}, it iteratively trains linear classifiers to predict a concept and projects the representations onto the intersection of their null spaces: $h_{k} = \left(\prod_{i=1}^k P_{N(w_i)}\right) h$, where $P_{N(w_i)}$ is the projection matrix onto the null space of the $i$-th classifier.
    
    \item \textbf{Truncated Polynomial Classifier (TPC):} Increases expressivity by adding controlled 2nd-degree (quadratic) pairwise feature interactions \citep{beyondlinearprobes}. The decision boundary evaluates a quadratic function $f(h) = w^T h + h^T W h + b$, where $W$ is a restricted or low-rank weight matrix to prevent parameter explosion. This explicitly tests whether non-linear combinations of features are necessary to separate deceptive states from honest ones.
    
    \item \textbf{Follow-Up Probe:} An active-probing operational baseline. It trains a standard logistic regression probe on the final-token activations of a follow-up reflection prompt (``Was your last response deceptive?'') with the answer prefilled as ``No'' \citep{apollo}. 
    
    \item \textbf{Upper-Bound Probe:} A dataset-specific ceiling on probe performance. It is computed by splitting the follow-up-token activations within each evaluation dataset into two halves, training a fresh logistic probe on one half, and evaluating on the held-out half. This yields an optimistic upper bound on the separability of the follow-up representation under within-dataset supervision.
\end{itemize}

\subsection{Dense Versus Sparse Pipelines}
To isolate the effects of representation types, we evaluate \texttt{Gemma-2-27B} on dense activations and \textit{Gemma Scope} \citep{gemmascope} on sparse SAE features. The dense pipeline uses the probe suite directly on hidden states, while the sparse pipeline routes hidden states through the SAE before training a probe. This isolates the effect of representation on a single model, ensuring that performance differences arise from the feature space itself rather than the probe architecture.

\section{Experiments}

\begin{table*}[h]
\centering
\caption{A strictly controlled comparison of dense hidden states and sparse SAE features extracted from the exact same foundation model (\texttt{Gemma-2-27B}) at 66th percentile layer. By evaluating both representation types on the identical underlying model, we isolate the specific effect of sparsity on detectability. Both variants were trained on the full DolusChat mixture and evaluated across CG, GS, HP-C, and HP-KR.}
\label{tab:sparsity_comparison}
\begin{tabular}{@{}l | cc | cc | cc | cc | c@{}}
\toprule
\multirow{2}{*}{\textbf{Method}} & \multicolumn{2}{c|}{\textbf{CG}} & \multicolumn{2}{c|}{\textbf{GS}} & \multicolumn{2}{c|}{\textbf{HP-C}} & \multicolumn{2}{c|}{\textbf{HP-KR}} & \multirow{2}{*}{\textbf{Avg}} \\
\cmidrule(lr){2-3} \cmidrule(lr){4-5} \cmidrule(lr){6-7} \cmidrule(lr){8-9}
 & Dense & SAE & Dense & SAE & Dense & SAE & Dense & SAE & \\
\midrule
\multicolumn{10}{c}{\textbf{AUROC}} \\
\midrule
Logistic   & \textbf{0.67} & \textbf{0.67} & \textbf{0.51} & 0.46 & \textbf{0.48} & 0.48 & \textbf{0.35} & 0.31 & 0.49 \\
TPC        & \textbf{0.69} & 0.66 & \textbf{0.48} & 0.43 & 0.51 & \textbf{0.52} & \textbf{0.54} & 0.48 & \textbf{0.54} \\
Truth2D    & \textbf{0.66} & 0.65 & \textbf{0.46} & 0.44 & \textbf{0.48} & 0.47 & \textbf{0.38} & 0.34 & 0.49 \\
Mass-Mean  & 0.53 & \textbf{0.63} & 0.44 & \textbf{0.45} & 0.48 & \textbf{0.49} & \textbf{0.37} & 0.34 & 0.47 \\
INLP       & 0.67 & \textbf{0.68} & \textbf{0.50} & 0.45 & \textbf{0.48} & 0.47 & \textbf{0.36} & 0.30 & 0.49 \\
Follow-Up  & \textbf{0.48} & 0.46 & \textbf{0.58} & 0.56 & 0.53 & \textbf{0.53} & 0.43 & \textbf{0.41} & 0.50 \\
Upper-Bound& \textbf{0.92} & \textbf{0.92} & 0.77 & \textbf{0.78} & \textbf{0.61} & \textbf{0.61} & \textbf{0.79} & 0.76 & 0.77 \\
\midrule
\textit{Average} & 0.66 & \textbf{0.67} & \textbf{0.53} & 0.51 & \textbf{0.51} & \textbf{0.51} & \textbf{0.46} & 0.42 & 0.53 \\
\midrule
\multicolumn{10}{c}{\textbf{RECALL}} \\
\midrule
Logistic   & \textbf{0.93} & 0.86 & \textbf{0.22} & 0.11 & \textbf{0.32} & 0.29 & \textbf{0.32} & 0.13 & 0.40 \\
TPC        & \textbf{0.84} & 0.78 & 0.20 & \textbf{0.16} & 0.38 & \textbf{0.40} & 0.20 & \textbf{0.36} & \textbf{0.42} \\
Truth2D    & \textbf{0.90} & 0.87 & 0.11 & \textbf{0.11} & \textbf{0.31} & \textbf{0.31} & \textbf{0.24} & 0.13 & 0.37 \\
Mass-Mean  & 0.19 & \textbf{0.66} & \textbf{0.37} & 0.18 & \textbf{0.47} & 0.41 & \textbf{0.46} & 0.28 & 0.38 \\
INLP       & \textbf{0.92} & 0.79 & \textbf{0.12} & 0.03 & \textbf{0.28} & 0.23 & \textbf{0.22} & 0.10 & 0.34 \\
Follow-Up  & 0.57 & \textbf{0.61} & 0.47 & \textbf{0.48} & \textbf{0.20} & 0.19 & 0.32 & \textbf{0.38} & 0.40 \\
Upper-Bound& 0.82 & \textbf{0.84} & 0.70 & \textbf{0.77} & \textbf{0.40} & 0.32 & \textbf{0.59} & 0.54 & 0.62 \\
\midrule
\textit{Average} & 0.74 & \textbf{0.77} & \textbf{0.31} & 0.26 & \textbf{0.34} & 0.31 & \textbf{0.33} & 0.27 & 0.42 \\
\bottomrule
\end{tabular}
\end{table*}

\subsection{Experimental Setup}
To prevent confounding variables and ensure rigorous evaluation, we divide our experiments into two distinct testing regimes. 

For the general depth and expressivity evaluations (Table~\ref{tab:layers_comparison}), we extract dense activations from \texttt{Mistral-Small-3.1-24B} at two depths: an early 20th-percentile baseline and a deeper 66th-percentile layer. 

For the sparsity analysis (Table~\ref{tab:sparsity_comparison}), we establish a strictly controlled comparison by utilizing \texttt{Gemma-2-27B} for \textit{both} the dense and sparse pipelines. This guarantees that any observed performance differences arise solely from the feature representation rather than a mismatch in the underlying foundation model. We extract both the raw dense hidden states and the corresponding sparse SAE features (using the \texttt{gemma-scope-27b-pt-res-canonical} weights with a hidden dimension of 131,072) from Layer 31. Because \texttt{Gemma-2-27B} features exactly 46 layers, this target corresponds to approximately 67.4\% depth, aligning with the peaked middle-layer activation pattern observed in \citet{caughtintheact}. 

Across all setups, training utilizes 1,000 DolusChat samples, a learning rate of $10^{-3}$, 20 probe epochs, and a batch size of 64.

\subsection{Evaluation Metrics}
To assess the detectability of deception across different representations, we report the Area Under the Receiver Operating Characteristic Curve (AUROC) and standard Recall (evaluated at a default 0.5 probability threshold). AUROC serves as our primary metric, because it provides a threshold-independent measure of the global linear separability between truthful and deceptive internal states. Rather than optimizing for a specific operational threshold, this approach allows us to answer our core mechanistic question: to what extent do dense versus sparse representations inherently encode the structural typology of a lie?

We acknowledge that in real-world operational deployments, lie detectors require strict False Positive Rate (FPR) constraints or precision-focused tuning, as over-predicting deception can falsely penalize honest model behavior. Because our primary objective is to evaluate fundamental representation geometries rather than to deploy a production-ready classifier, we leave the exact mapping of FPR trade-offs at strict operational thresholds to future work.

\section{Results and Analysis}
In this section, we present the empirical findings from our probing experiments. To systematically unpack the mechanics of deception detection in large language models, we structure our analysis around the four core factors hypothesized in Section 1: representation depth, probe expressivity, sparsity, and lie typology. For each dimension, we present our quantitative evaluations followed by an interpretation of the underlying mechanisms driving these results.

 \begin{figure*}[t]
    \centering
    \includegraphics[width=1\textwidth]{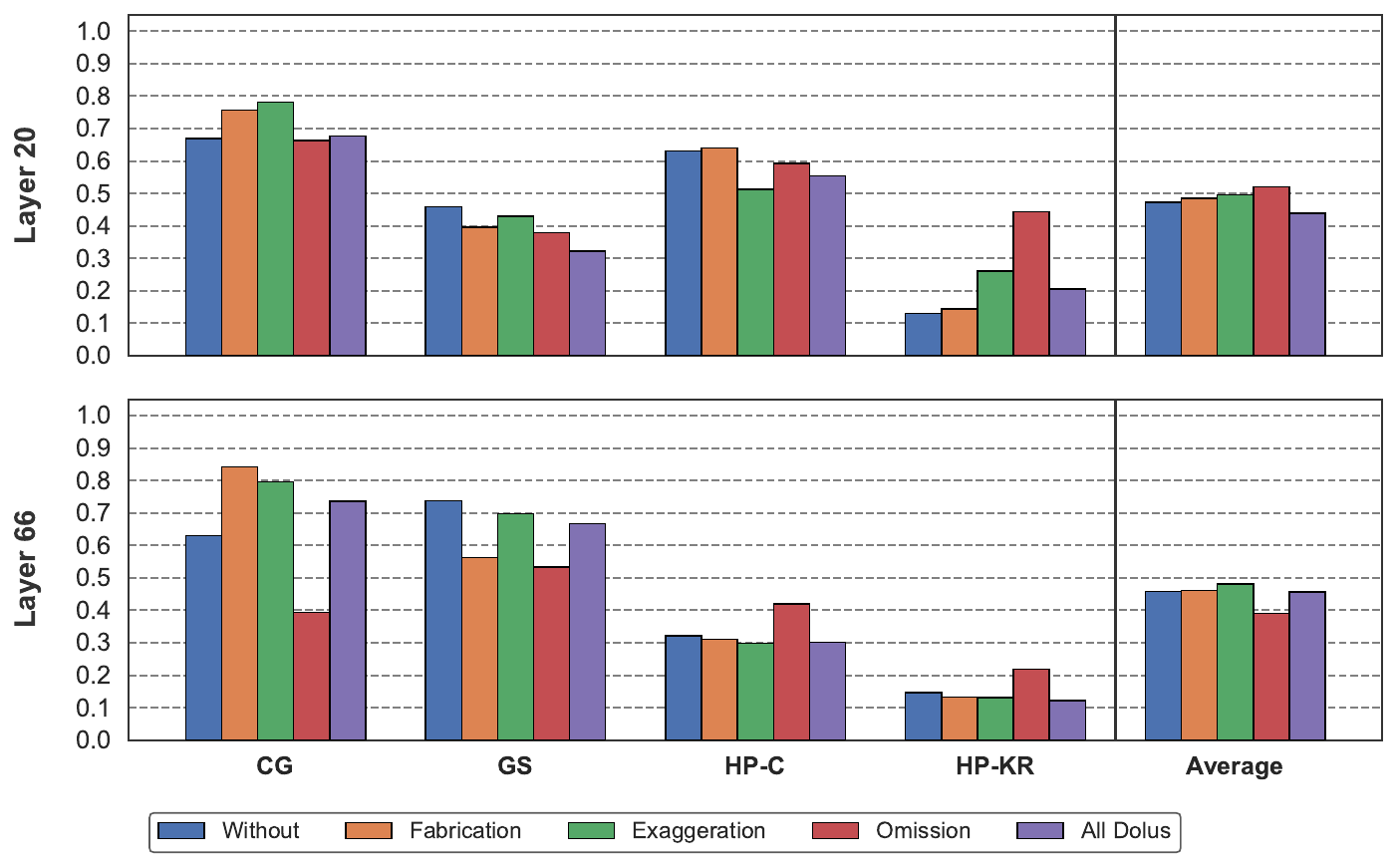}
    \caption{Comparison of logistic-probe AUROC across lie-typology training conditions and evaluation datasets. We compare probes trained without DolusChat, on fabrication-only DolusChat, on exaggeration-only DolusChat, on omission-only DolusChat, and on the full DolusChat mixture, at 20th-percentile and 66th-percentile.}
    \Description{Bar chart comparing logistic-probe AUROC across multiple lie-typology conditions including fabrication, omission, and exaggeration over the four evaluation datasets, showing the relative performance at both 20th-percentile and 66th-percentile.}
    \label{fig:topology_logistic}
\end{figure*}

\subsection{The Effect of Representation Depth}

To investigate whether deceptive representations mature in the later stages of the model, we extracted dense hidden states at two distinct depths: an early layer (the 20th percentile) and a deeper layer (the 66th percentile). As shown in Table~\ref{tab:layers_comparison}, the effect of representation depth is highly dataset-dependent rather than universally monotonic.

On the self-referential datasets (CG and GS), the 66th percentile layer consistently improves AUROC across most probes. This suggests that for role-based or secret-keeping deception, deeper representations are needed to capture the internal contradiction between the model's assigned persona and its underlying knowledge.

Conversely, the harm-pressure datasets (HP-C and HP-KR) exhibit stronger separability at the 20th percentile layer. This early-layer advantage possibly occurs because deceptive responses in these datasets are heavily entangled with safety protocols, refusal tendencies, and harmful-context cues. In these cases, deeper layers may over-abstract the representation, washing out surface-level features that linear probes rely on to detect the lie.

Furthermore, the recall metrics reveal a distinct split depending on the dataset category. On the harm-pressure datasets, the 20th percentile layer provides a more robust and stable decision boundary at the default operating threshold. Conversely, for the baseline logistic probe, the 66th percentile layer yields higher recall on the self-referential datasets (CG and GS). This reinforces the broader conclusion that the optimal representation depth is strictly tied to the nature of the deceptive context rather than a universal property of the model.

A striking anomaly in Table~\ref{tab:layers_comparison} is the sub-random AUROC performance (e.g., 0.12--0.16) on the HP-KR dataset across nearly all probes. It is important to note that this is not an erroneous result; the original Liars' Bench benchmark also reports substantially sub-chance AUROC results in this setting (for example, finding an AUROC of 0.12 for Llama 3.3 70B with a mean probe on HP-KR) \citep{liarsbench}. Because an AUROC significantly below 0.50 indicates systematic anti-correlation, this is not merely a failure to generalize, but an active \textit{anti-transfer}. The probes trained on direct, context-free lies in DolusChat appear to latch onto a feature direction that maps inversely to the safety-driven, self-knowledge deception required in HP-KR. This is a crucial mechanistic insight: the internal representation of a direct fabrication is inverted compared to the representation of a model falsely denying hazardous knowledge under safety pressure.

\subsection{Impact of Probe Expressivity}
To investigate whether more expressive probe families---those capturing feature interactions or subspace structure---outperform standard logistic regression, we compare logistic regression against TPC, Truth2D, Mass-Mean, and INLP. We treat the follow-up and upper-bound rows separately, since they serve as prompt-based and ceiling baselines rather than expressivity-oriented methods.

As shown in Table~\ref{tab:layers_comparison}, the empirical results provide mixed support for the expressivity hypothesis. Truth2D provides the clearest gains among the structured probes, particularly on GS, where it slightly improves AUROC over the logistic regression baseline. This is consistent with the idea that explicitly disentangling truth from polarity can create a more robust deception subspace. Conversely, TPC does not yield a consistent improvement over logistic regression, suggesting that low-order polynomial interactions are either insufficient for this task or too sensitive to the training distribution to generalize reliably. Mass-Mean is generally the weakest of the structured probes, indicating that this simplified geometric summary does not preserve enough discriminative information.

Crucially, while INLP serves as a useful geometric baseline, its performance should be interpreted as a measure of subspace robustness rather than standard detection. As shown in Table~\ref{tab:layers_comparison}, INLP is often close to logistic regression, but not identical across all datasets and depths: on CG it reaches 0.74 and 0.82 AUROC versus 0.76 and 0.84 for logistic regression, on GS it reaches 0.41 and 0.55 versus 0.40 and 0.56, on HP-C it matches the baseline at 0.64 and 0.31, and on HP-KR both methods remain near chance. These results suggest that iteratively projecting out linear directions removes some information, but does not uniformly collapse the deception signal. Rather than indicating that deception is confined to a single low-dimensional subspace, the findings are more consistent with a partly distributed representation that is only partially removed by linear null-space projection.

\subsection{Impact of Sparse SAE Features}

Unlike the earlier fabrication-only comparison, this experiment uses SAE results computed across all DolusChat lie categories. Furthermore, due to computational constraints, SAE features were only extracted at 66th percentile layer. To keep the comparison matched, we therefore evaluate the dense branch on the same all-category DolusChat subset and at the same layer, so any differences reflect the representation type rather than changes in lie-type coverage or depth.

To examine whether sparse SAE features preserve or improve deception separability relative to dense hidden states, we compare Gemma-2-27B hidden activations against Gemma-2-27B SAE features at 66th percentile layer in Table~\ref{tab:sparsity_comparison}. 

The results presented in the table suggest that, within this specific experimental setup, the transition from dense hidden states to sparse SAE features does not appear to yield a clear improvement in deception detection performance. While the sparse representations maintain a relative functional parity with the dense branch, they generally achieve slightly lower or equivalent average scores across most datasets and probing methods. This suggests that the SAE reconstruction at this depth preserves existing deception-related signals rather than enhancing their separability.

The main exception is the harm-pressure setting, where SAE features improve some individual probe scores and appear to help disentangle deception from overlapping safety-related signals. However, this effect is not uniform across all methods or datasets, and it does not translate into a consistent overall gain. The upper-bound rows further support this interpretation: dense and sparse features remain close in performance, but the dense branch is still slightly stronger overall.

\subsection{Impact of Lie Typology}

To test whether the structural category of a lie affects detectability, we compare AUROC across five training conditions: without DolusChat, fabrication-only, exaggeration-only, omission-only, and the full DolusChat mixture. We restrict this analysis to the baseline logistic regression probe because, as established in Section 5.2, it provides the most stable and representative baseline across datasets. By standardizing on this robust linear boundary, we successfully isolate the effect of the training distribution on linear separability without compounding variables from fluctuating probe expressivity. Figure~\ref{fig:topology_logistic} summarizes these results at the 20th percentile layer and the 66th percentile layer.

The bar plot shows that lie typology matters, but not in a simple universal order. At the 20th percentile, the best-performing condition depends on the dataset: exaggeration is strongest on CG, the no-DolusChat baseline is strongest on GS, fabrication is strongest on HP-C, and omission is strongest on HP-KR. Averaged across datasets, omission has the highest AUROC at the 20th percentile layer, largely because of its strong performance on HP-KR. This suggests that omission is the most recoverable training condition at the shallower layer, but only in the harm-pressure setting.

At the 66th percentile layer, the pattern shifts. Fabrication becomes strongest on CG, the no-DolusChat baseline remains strongest on GS, and omission becomes the strongest condition on both HP-C and HP-KR. On average, exaggeration achieves the highest AUROC at the 66th percentile layer, while omission drops substantially relative to its 20th percentile layer performance. This indicates that deeper representations sharpen some lie-typology signals while suppressing others, especially when the lie depends on contextual withholding rather than direct contradiction.

The full DolusChat mixture is never the best-performing condition and is weakest at the 20th percentile layer, with a near-bottom average at the 66th percentile layer as well. This suggests that combining lie types does not automatically improve linear separability and may instead blur the deception signal. Overall, the figure supports a qualified conclusion: lie typology does affect detectability, but there is no single ordering of fabrication, exaggeration, and omission that holds across all datasets or depths.

\paragraph{Summary: A Representation-Dependent Problem.} The results point to a consistent but nuanced picture of deception detection in LLMs. Depth improves separability in some datasets but not others, so later layers are not universally superior. More expressive probes help selectively, but standard logistic regression remains highly competitive and is often the most stable baseline. Sparse SAE features perform similarly to dense representations. Finally, lie typology clearly matters: fabrication, omission, exaggeration, and mixed training conditions produce different detection patterns depending on the dataset and representation layer. Overall, the experiments show that deception detection is best understood as a representation-dependent problem rather than a single-classifier problem.

\section{Conclusion}
This paper set out to examine deception detection in large language models through four factors: representation depth, probe expressivity, sparse representations, and lie typology. Across the experiments, the central conclusion is that deception is not encoded in a single uniform way. Instead, detectability depends heavily on the dataset context, the layer being probed, and the representation space used for classification.

For depth, the results show a dataset-dependent pattern rather than a universal rule. Some datasets benefit from deeper layers, while others remain more separable at earlier layers, meaning the commonly assumed ``later is better'' trajectory does not hold consistently across all deception settings. For probe expressivity, more complex models provide selective gains but do not dominate the simple logistic baseline. Truth2D offers the clearest structured improvement, while TPC, mass-mean, and INLP are useful in specific settings but not consistently stronger than a well-calibrated linear classifier. This suggests that deception detection is not solved simply by adding geometric or polynomial complexity.

For sparse SAE features, the results suggest that evaluating SAE representations at 66th percentile layer does not consistently enhance detectability over the original dense hidden states. In most settings, SAE preserves the core deceptive signal rather than improving it, and its benefits appear limited to a subset of harm-pressure cases. This suggests that, at least at this depth, SAE features function more as an alternative representation than as a reliably stronger one for deception detection. Finally, for lie typology, we found that fabrication, omission, exaggeration, and mixed training conditions do not behave uniformly. The strongest condition depends on both the dataset and the representation layer, indicating that the structural form of a lie materially affects its detectability, and no single lie type dominates universally.

The main challenge throughout this paper was that the deception signal was not stable across tasks; different datasets exposed different failure modes, and the optimal probing strategy frequently shifted with the evaluation context. Given that optimal detection performance depends heavily on highly localized factors---such as specific layers, lie typologies, and representation spaces---our findings strongly suggest that a monolithic detector is fundamentally limited. Instead, a highly promising avenue for future work lies in ensemble methods or a Mixture-of-Experts (MoE) probing framework. By training specialized, expert probes for distinct context types (e.g., separating safety-related choices from persona-driven roleplay) and utilizing a gating mechanism to dynamically route activations, such an architecture could explicitly leverage the selective strengths of the individual probes observed in our evaluation. Additionally, extending this analysis to the remaining Liars' Bench models would test whether these depth, sparsity, and typology patterns persist across different model families, while studying the SAE representations more directly could further bridge the gap between behavioral benchmarking and mechanistic interpretability.

\section*{Limitations}
This study has several limitations. First, given the computational scope of a multi-factor study and the constraints of an extended abstract, we sampled two specific depth percentiles (20\% and 66\%) rather than conducting a continuous layer-wise sweep. Because we purposefully anchored our early layer to the original Liars' Bench baseline and our deeper layer to the maximum theoretical peak before the late-stage performance drop-off identified by \citet{caughtintheact}, we chose to apply these established bounds directly to our evaluation rather than exhaustively re-proving a continuous layer-wise trajectory from scratch. Second, because our core claims are intentionally conservative---highlighting contextual dependencies (``it depends'') rather than declaring absolute algorithmic superiority---we rely on directional shifts in AUROC and refrain from exhaustive statistical significance testing. Third, the lie typology experiments (Figure~\ref{fig:topology_logistic}) were restricted to the logistic regression baseline to cleanly isolate distribution effects, and have not yet been evaluated across the more expressive probe families. Fourth, the SAE results were evaluated solely at 66th percentile layer. Fifth, the inherent distribution gap between our synthetic training data (DolusChat) and the complex evaluation prompts (Liars' Bench) likely contributes to the striking anti-transfer observed in the HP datasets. Finally, our discrete classification metrics rely on a default 0.5 probability threshold, whereas future operational work must explore strict False Positive Rate (FPR) constraints.

\section*{Acknowledgments}

This research was supported by the state of North Rhine-Westphalia as part of the Lamarr Institute for Machine Learning and Artificial Intelligence.

\bibliographystyle{ACM-Reference-Format}
\bibliography{references}

\end{document}